\documentclass{article}

\usepackage[preprint]{neurips_2024}
\usepackage{lipsum}
\usepackage{comment}
\usepackage{multirow}
\usepackage{caption}
\usepackage{rotating}
\usepackage{scalerel}
\usepackage{amsmath}
\usepackage{multicol}
\usepackage{blindtext}
\usepackage{algpseudocode}
\usepackage[table,xcdraw]{xcolor}

\usepackage{pifont}
\newcommand{\cmark}{\ding{51}}%
\newcommand{\xmark}{\ding{55}}%

\usepackage{enumitem}

\usepackage[utf8]{inputenc} 
\usepackage[T1]{fontenc}    
\usepackage{hyperref}       
\usepackage{url}            
\usepackage{booktabs}       
\usepackage{amsfonts}       
\usepackage{nicefrac}       
\usepackage{microtype}      
\usepackage{xcolor}         
\usepackage{bigfoot}

\newcommand{\model}{\textsc{PixArt-$\alpha$}\xspace}

\hypersetup{
    colorlinks=true,
    linkcolor=green, 
    citecolor=green, 
    urlcolor=blue    
}

\title{FORA: Fast-Forward Caching in Diffusion Transformer Acceleration}

\author{%
Pratheba Selvaraju$^1$\\
\And
Tianyu Ding$^{2}$\thanks{Corresponding author. Preprint version.} \\
\And
Tianyi Chen$^2$\\
\And
Ilya Zharkov$^2$\\
\And
Luming Liang$^2$
}

\date{{$^1$UMass Amherst  \qquad  $^2$Microsoft}\\
{\tt\small pselvaraju@cs.umass.edu},
{\tt\small \{tianyuding,tiachen,zharkov,lulian\}@microsoft.com}
}

\begin{document}

\maketitle

\begin{figure}[ht!]
    \centering
\includegraphics[width=.98\linewidth]{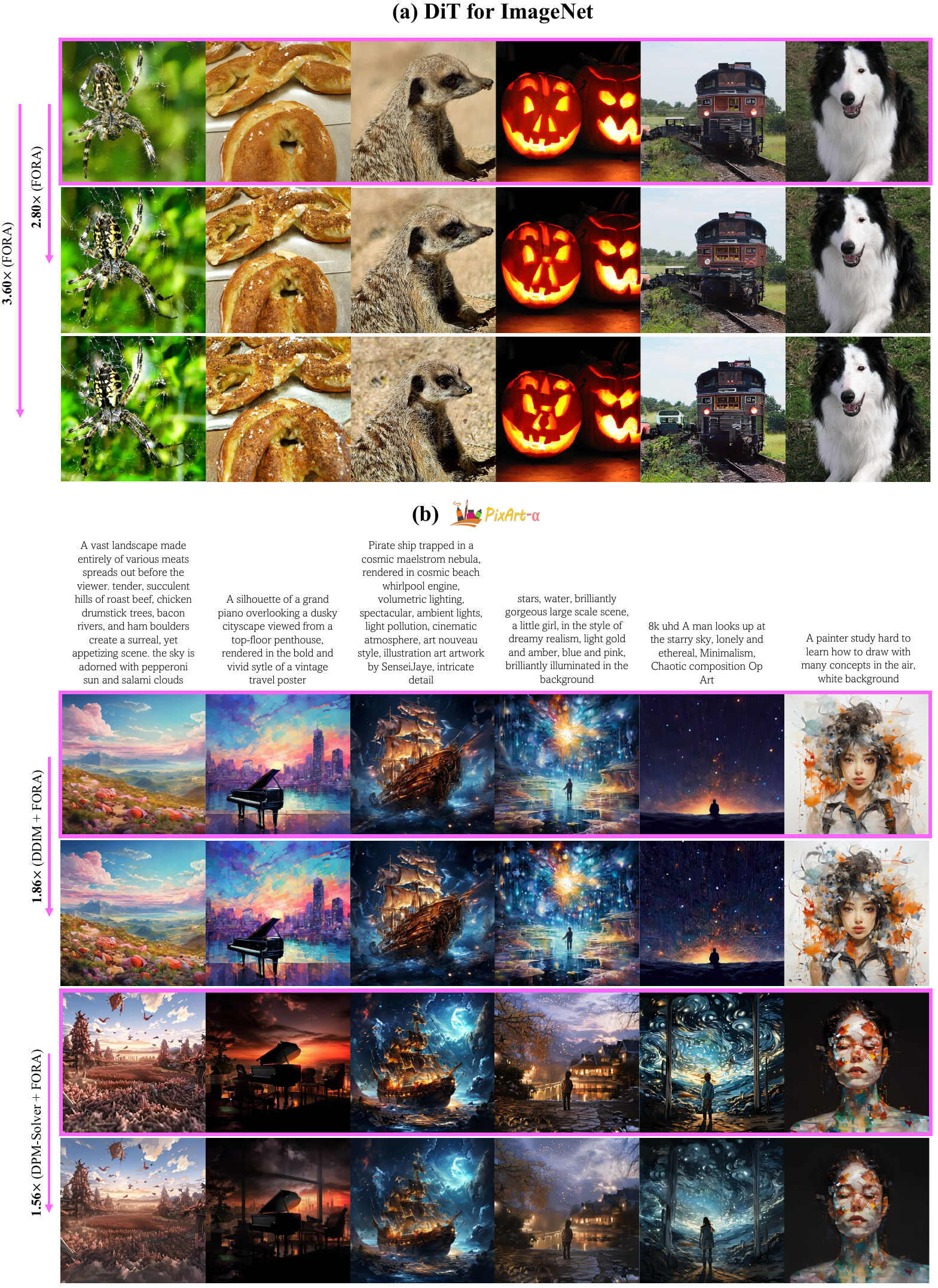}
    \caption{Image generations with FORA on DiT  and \model. The image sizes are 512 $\times$ 512.
    }
    \label{fig:teaser}
    \vspace{-.19in}
\end{figure}

\begin{abstract}
  Diffusion transformers (DiT) have become the de facto choice for generating high-quality images and videos, largely due to their scalability, which enables the construction of larger models for enhanced performance. However, the increased size of these models leads to higher inference costs, making them less attractive for real-time applications. We present \textbf{F}ast-F\textbf{OR}ward C\textbf{A}ching (FORA), a simple yet effective approach designed to accelerate DiT by exploiting the repetitive nature of the diffusion process. FORA implements a caching mechanism that stores and reuses intermediate outputs from the attention and MLP layers across denoising steps, thereby reducing computational overhead. This approach does not require model retraining and seamlessly integrates with existing transformer-based diffusion models. Experiments show that FORA can speed up diffusion transformers several times over while only minimally affecting performance metrics such as the IS Score and FID. By enabling faster processing with minimal trade-offs in quality, FORA represents a significant advancement in deploying diffusion transformers for real-time applications. Code will be made publicly available at: \url{https://github.com/prathebaselva/FORA}.
\end{abstract}

\section{Introduction}
\label{sec:intro}

Diffusion models~\citep{sohl2015deep,ho2020denoising} have gained significant attention in generative tasks due to their exceptional ability to produce high-quality and diverse outputs. Initially conceptualized with the U-Net architecture, these models have demonstrated impressive performance in domains such as image and video generation~\citep{rombach2022high,ho2022video,saharia2022photorealistic}. However, the scalability of U-Net-based diffusion models is inherently limited, posing challenges for applications that require larger model capacities to achieve superior performance. To overcome these, diffusion transformers (DiT)~\citep{peebles2023dit} have emerged as the preferred choice for generative tasks. By leveraging the inherently scalable architecture of transformers, DiT models facilitate effective scaling of model capacities. This scalability has enabled the development of larger models, resulting in improved performance in generating high-quality images and videos. Despite their advantages, the increased size of DiT models leads to higher inference costs, making them less suitable for real-time applications.

Recently, a new line of research has focused on accelerating diffusion models during inference by recognizing the iterative nature of the  sampling process~\citep{ma2023deepcache,wimbauer2023cachemeifyoucan,agarwal2024approximate,zhang2024cross,yu2024accelerating}. These caching mechanisms exploit the repetitive nature of the diffusion process, storing and reusing intermediate outputs to reduce computational overhead. However, existing caching-based efficiency methods primarily target U-Net-based diffusion models and do not specifically address transformer-based models.

This paper aims to bridge this gap by focusing on transformer-based diffusion models. We introduce Fast-Forward Caching (FORA), a simple yet effective approach designed to accelerate DiT models by leveraging the repetitive nature of the diffusion process. FORA implements a caching mechanism that stores and reuses intermediate outputs from the attention and MLP layers across denoising steps, thereby reducing computational overhead. This approach does not require model retraining and seamlessly integrates with existing transformer-based diffusion models. By enabling faster processing with minimal trade-offs in quality, FORA represents a significant advancement in deploying diffusion transformers for real-time applications.

In summary, we make the following specific contributions:
\begin{itemize}[leftmargin=*]
    \item We propose Fast-Forward Caching (FORA), a caching strategy tailored for transformer-based diffusion models. This mechanism capitalizes on the repetitive aspects of the diffusion process by preserving and reusing intermediate outputs from attention and MLP layers during inference. 

    \item FORA significantly cuts down computational overhead and integrates seamlessly with existing DiT models without necessitating retraining. As a result, it effectively reduces computational costs while maintaining output quality.

    \item We perform experiments to assess the performance of FORA, demonstrating notable improvements in inference speed and computational efficiency. Our findings underscore FORA's potential to make high-performance generative models more suitable for real-time use.
\end{itemize}

\section{Related work} \label{sec:related work}

\subsection{Diffusion model}

In the realm of generative models, diffusion models~\citep{ho2020denoising,sohl2015deep} have emerged as a cornerstone for their remarkable ability to generate high-quality and diverse outputs. Initially conceptualized with the U-Net architecture, these models have shown impressive performance in areas such as image and video generation~\citep{rombach2022high,ho2022video,saharia2022photorealistic}. Despite their success, the scalability of U-Net-based diffusion models is inherently limited, presenting challenges for applications that demand larger model capacities for improved performance. In addressing this, the advent of diffusion transformers (DiT)~\citep{peebles2023dit} marks a significant step forward. Leveraging the inherently scalable architecture of transformers~\citep{vaswani2017attention}, DiT offers a pathway to effectively scale up model capacities. A notable achievement in this direction is the breakthrough in generating long videos through the large-scale training of Sora~\citep{videoworldsimulators2024}, which employs this transformer-based diffusion architecture for general-purpose simulations of the physical world. This highlights the significant impact of scaling transformer-based diffusion models.

However, the escalation in scale and capacity brings forth a crucial challenge: the extensive computation necessitated by large-scale transformer-based diffusion models. Specifically, the computational needs of these models, particularly concerning inference speed and resource consumption, pose a bottleneck for real-time applications and efficient deployment.  This paper aims to address this pivotal issue by improving the inference efficiency of transformer-based diffusion models in a training-free manner, all while preserving their generative capabilities.

\subsection{Efficiency enhancements of diffusion model}

Recent research has focused on various strategies to enhance the efficiency of diffusion models, addressing both training and inference costs.

\textbf{Training Efficiency.} Training efficiency refers to the ability to reduce the computational resources, time, and energy required to train models without compromising their performance. Efficient training methods are critical for making advanced diffusion models more accessible and environmentally sustainable. One notable advancement in this area is \textsc{PixArt} series~\citep{chen2023pixartalpha,chen2024pixart2,chen2024pixart}, which introduces a novel training strategy decomposition, which significantly reduces training costs and CO2 emissions while achieving competitive image generation quality compared to existing models. Additionally, DREAM~\cite{zhou2023dream} addresses the misalignment of training with sampling in diffusion models  by proposing a novel training framework that achieves several times faster training convergence.

\textbf{Inference Efficiency.} Inference efficiency focuses on accelerating the inference phase of diffusion models without compromising generative quality. Efficient techniques primarily involve reducing sampling steps~\citep{song2020denoising,bao2021analytic,liu2021pseudo,lu2022dpm,zheng2024dpm,song2023consistency,salimans2021progressive} and model compression~\citep{fang2024structural,he2024ptqd,shang2023post}. Recently, recognizing the iterative nature of the diffusion sampling process, a new line of research has emerged around caching mechanisms to accelerate diffusion models during inference.  DeepCache~\citep{ma2023deepcache} leverages temporal redundancy by caching and retrieving features across adjacent denoising stages, reducing redundant computations and speeding up inference. Block caching~\citep{wimbauer2023cachemeifyoucan} reuses outputs from layer blocks of previous steps to cut down on redundant computations, though it requires a lightweight training stage to address artifacts.  \cite{agarwal2024approximate} employs an approximate-caching technique that reuses intermediate noise states from previous image generations to skip initial denoising steps, achieving significant GPU compute savings, latency reduction, and cost savings. \cite{wang2024attention} prunes redundant tokens at runtime using attention maps. \cite{yu2024accelerating} is a cache-enabled sparse diffusion inference engine designed for text-to-image editing, identifying and updating only the affected regions of an image based on textual modifications.  \cite{zhang2024cross} caches and reuses cross-attention outputs once they converge, reducing computational complexity while maintaining performance. However, all these  caching-based efficiency methods focus on conventional U-Net-based diffusion models and do not specifically consider transformer-based models. In contrast, our approach addresses the acceleration of transformer-based diffusion models through a caching mechanism that is both training-free and plug-and-play, aligning with the growing popularity and recent advancements in these models.

\section{Fast-Forward Caching for Accelerated Sampling}
\label{sec:method}


\begin{figure}
    \centering
    \includegraphics[width=.98\linewidth]{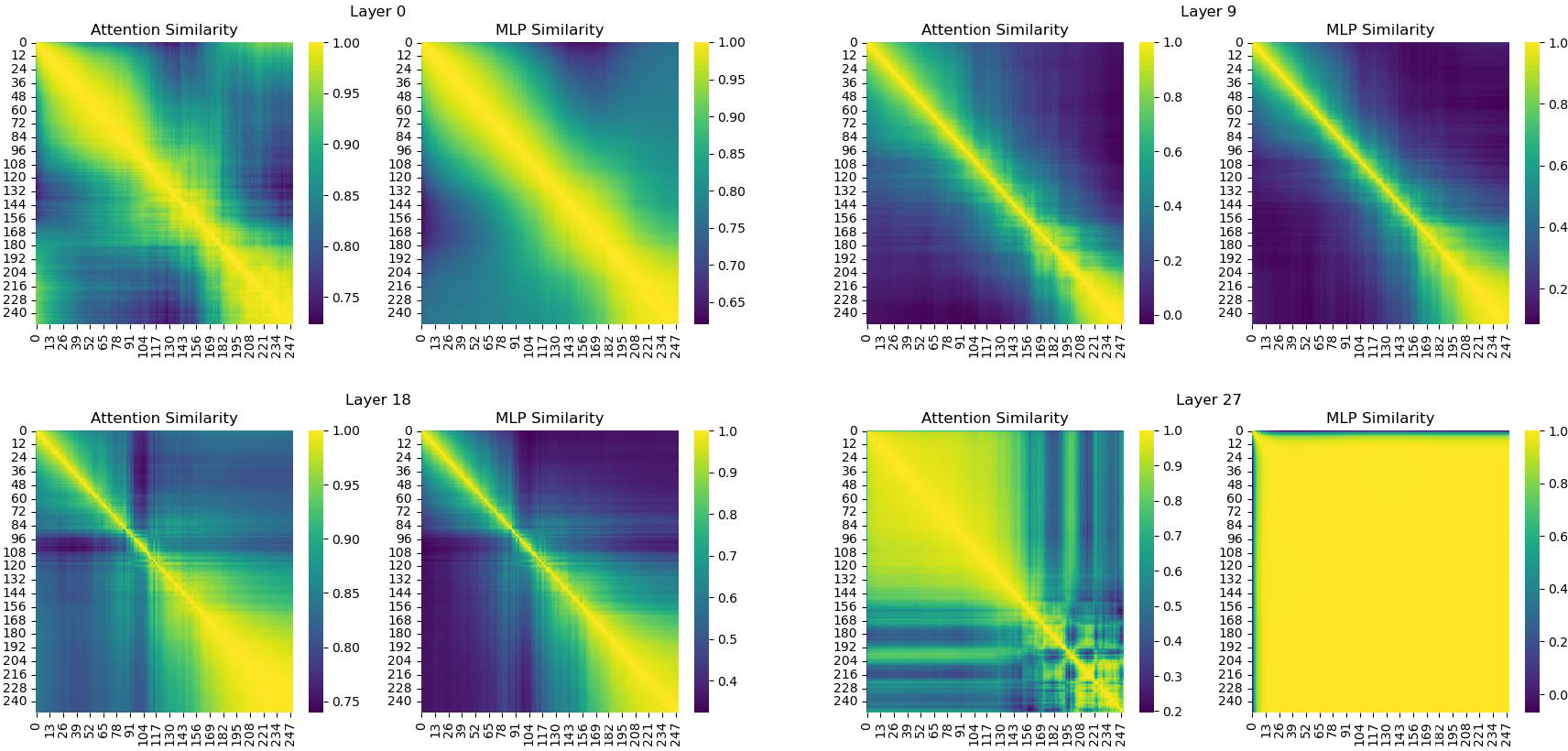}
    \caption{\textbf{Feature similarity analysis across time steps in DiT for attention and MLP layers.} This visualization is based on a 250-step DDIM sampling process. The heatmap illustrates the high degree of similarity between features of consecutive time steps, particularly in the later stages of the diffusion process. This observation motivates our caching strategy, highlighting potential areas for computational optimization without significant loss of information.}
    \label{fig:feature_similarity}
\end{figure}

Feature caching is a powerful technique that enhances speed and efficiency by storing and reusing computed information. This approach can reduce computational overhead and minimize latency. In this paper, we propose FORA (Fast-Forward Caching), which employs this technique to accelerate the sampling process in Diffusion Transformers (DiT) models.

\subsection{Motivations}

The development of FORA is grounded in a series of critical observations for DiT models. Our first key insight stems from the nature of the denoising process itself. As seen in Fig.~\ref{fig:feature_similarity}, we observe a striking visual similarity among the outputs of consecutive time steps during the sampling process. This similarity correlates strongly with a close feature similarity at the layer level across these consecutive time steps, opening up possibilities for computational efficiency.

Secondly, we recognize that the fundamental structure of the diffusion model remains constant throughout the sampling process. The noise removal objective and the overall sampling procedure maintain their integrity from the initial noisy state to the final generated image. This consistency in the model's behavior provides a stable foundation upon which we can build our caching strategies.

Our third crucial observation relates to the computational architecture of DiT models. We identify that the share of computational overhead is concentrated in specific components of the model, namely the self-attention and Multi-Layer Perceptron (MLP) layers. These layers, while essential for the model's performance, represent significant bottlenecks in terms of computational efficiency.

\subsection{FORA}

In light of the above observations, we design FORA to leverage the inherent similarities between consecutive time steps without compromising the model's underlying architecture. Our approach focuses on caching and reusing features from the computationally intensive layers mentioned above. By doing so, we can significantly reduce redundant calculations while maintaining the integrity of the generation process. This caching mechanism is informed by both qualitative assessments of visual similarity and quantitative measurements of feature similarity across time steps. In the following, we adopt upon DiT~\citep{peebles2023dit} as representative network and demonstrate the caching mechanism on it. However, our method is generalizable such that any similar attention or high computational overhead layers in a transformer backbone model can utilize our caching strategy. 

\begin{figure}[t]
    \centering
\includegraphics[width=1\linewidth]{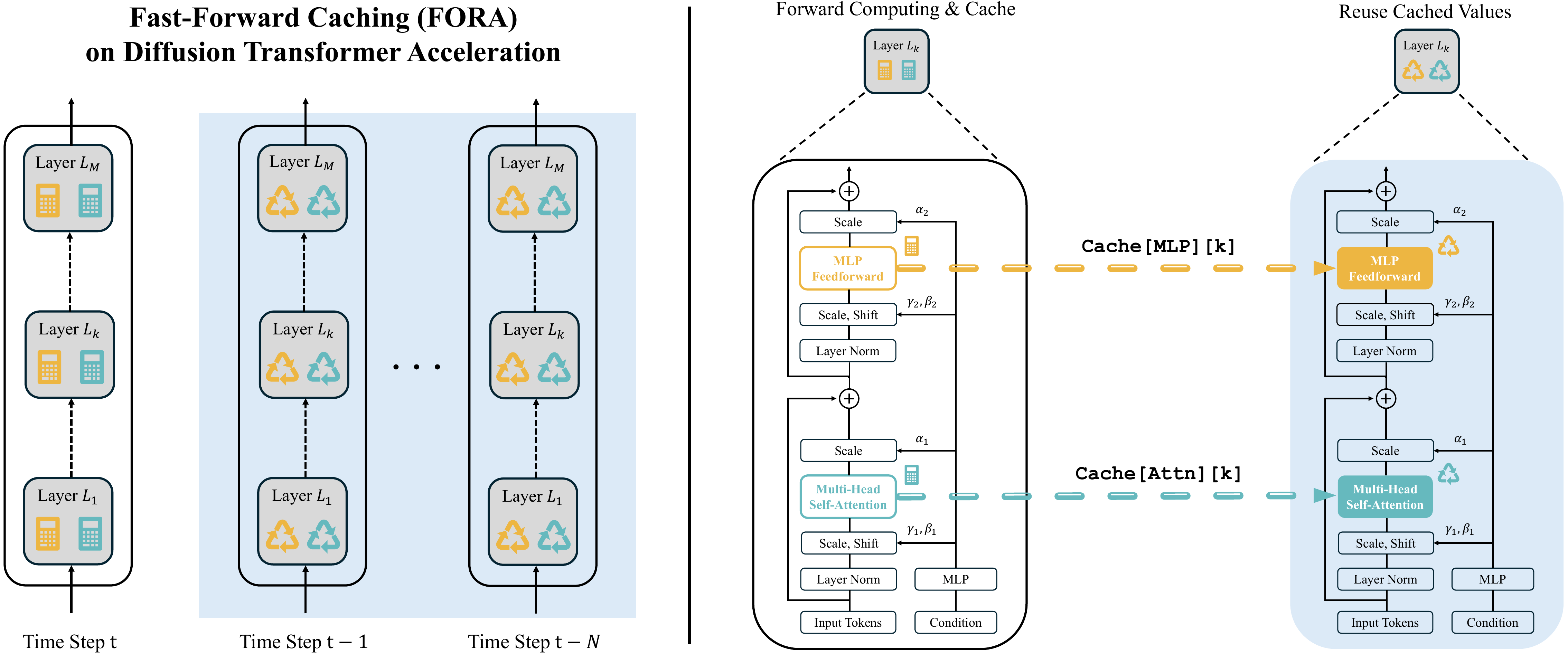}
    \caption{\textbf{Static Caching in FORA for DiT architecture}. The hyper-parameter $N$ determines the frequency of layer recomputation. At time step $t$, when $t \mod N = 0$, the model recomputes and caches features in \texttt{cache[Attn][k]} and \texttt{cache[MLP][k]} for each layer $k$. For subsequent time steps ($t-1$ to $t-N$), the model retrieves and reuses these cached features, avoiding redundant computations. This cycle repeats throughout the reverse diffusion process, balancing computational efficiency with output quality.}
    \label{fig:static_cache_mechanism}
\end{figure}

FORA implements a \emph{static caching} mechanism,  a straightforward yet powerful approach, to accelerate the sampling process in diffusion models. This method operates on a simple principle: recompute and cache features at regular intervals, and reuse these cached features for a predetermined number of subsequent time steps. At the core of this mechanism is a single hyperparameter $N$, which we call the \emph{cache interval}. This interval determines how frequently the model recomputes and caches new features. Specifically, $N$ is an integer that can range from 1 to $T-1$, where $T$ is the total number of sampling time steps in the diffusion process. The static caching process unfolds as follows:

\vspace{-.1in}
\begin{itemize}[leftmargin=*]
    \item \textbf{Initialization:} At the start of the sampling process ($t = T$), the model computes and caches the features for all layers.
    \item \textbf{Caching Condition:} For any given time step $t$, the model checks if $t$ is divisible by $N$ (i.e., $t \mod N = 0$). If this condition is met, it triggers a recomputation and caching event.
    \item \textbf{Recomputation and Caching:} When the caching condition is met, the model performs a full forward pass through all layers of the DiT block. The resulting features from the attention (\texttt{Attn}) and point-wise feed-forward (\texttt{MLP}) layers are then stored in the \texttt{cache} dictionary. Specifically, for each layer $k$, we store computed attention features in \texttt{cache[Attn][k]} and computed MLP features in \texttt{cache[MLP][k]}.
    \item \textbf{Feature Reuse:} For the subsequent $N-1$ time steps (i.e., until the next caching event), the model reuses the cached features instead of recomputing them. This means that for any layer $k$ during these steps, the model retrieves the features from \texttt{cache[Attn][k]} and \texttt{cache[MLP][k]} rather than performing the computationally expensive forward pass.
    \item \textbf{Cycle Repetition:} This process of recomputation, caching, and reuse continues cyclically until the sampling process completes at $t = 0$.
\end{itemize}

\subsection{Discussion}

The effectiveness of static caching hinges on the choice of the cache interval $N$. A smaller $N$ leads to more frequent recomputations, potentially preserving more accuracy but offering less computational savings. Conversely, a larger $N$ increases computational efficiency but may impact the quality of the generated outputs. In our experiments, we found that the optimal value of $N$ depends on the specific requirements of the task and the desired trade-off between speed and quality. Through extensive testing, we determined that setting $N$ max to 7 provides a good balance. Beyond this value, we observed a significant degradation in the Fréchet Inception Distance (FID) score, indicating a decline in the quality of generated images.

It's worth noting that while static caching is remarkably effective in reducing computational overhead, it does so in a uniform manner across all time steps. This approach, while simple to implement and tune, may not fully capitalize on the varying degrees of similarity between features at different stages of the denoising process. This observation led us to explore more dynamic caching strategies, which we leave as future iteration of the research.

\section{Experiments}
\label{sec:experiment}

In this section, we present a comprehensive overview of our experimental setup, including the datasets and evaluation models used to validate our method. We also delve into the details of our experimental configuration, evaluation metrics, and results, providing comparisons with baseline methods.

\begin{table*}[t]
  \centering
      \small
      \resizebox{\linewidth}{!}{
      \begin{tabular}{l | c c | c c c c c }
        \toprule
        \multicolumn{8}{c}{\bf ImageNet 256 $\times$ 256}  \\
        \bf Method & \bf Speed $\uparrow$ & \bf Retrain & \bf FID $\downarrow$ & \bf sFID $\downarrow$ & \bf IS $\uparrow$& \bf Precision $\uparrow$& \bf Recall $\uparrow$  \\
        \midrule
        IDDPM & - & \xmark &  12.26 & 5.42 & - & 0.70 & 0.62 \\
        \midrule
        Spectral DPM & - & \cmark  & 10.60 & - & - & - & - \\
        Diff-Pruning* &  - & \cmark & 9.27{\tiny(9.16)} & 10.59 & 214.42{\tiny (201.81)} & \textbf{0.87} & 0.30 \\  
      \midrule
       ADM~\citep{dhariwal2021adm} & & &10.94 & 6.02 & 100.98 & 0.69 & 0.63 \\
       ADM-U & & &7.49 & 5.13 & 127.49 & 0.72 & 0.63 \\
        ADM-G & & &4.59 & 5.25 & 186.70 & 0.82 & 0.52 \\
         ADM-G, ADM-U & & &3.94 & 6.14 & 215.84 & 0.83 & 0.53 \\
      \midrule
      CDM~\citep{ho2021cascaded} & & &4.88 & - & 158.71 & - & - \\
      \midrule
      LDM-8~\citep{rombach2022high} & & &15.51 & - & 79.03 & 0.65 & 0.63 \\
      LDM-8-G & & &7.76 & - & 209.52 & 0.84 & 0.35 \\
      LDM-4-G (cfg $=1.25$) & & &3.95 & - & 178.22 & 0.81 & 0.55 \\
      LDM-4-G (cfg $=1.50$) & & &3.60 & - & 247.67 & \textbf{0.87} & 0.48 \\
      \midrule
      {DiT-XL/2}     &  & \xmark& 9.62 & 6.85 & 121.50 & 0.67 & \textbf{0.67} \\
      {DiT-XL/2-G} (cfg $=1.25$) & & \xmark &3.22 & 5.28 & 201.77 & 0.76 & 0.62 \\
      {\textbf{DiT-XL/2-G} (cfg $=1.50$)} & 1$\times$ & {\xmark}& {2.27} & {4.60} & {278.24} & {0.83} & {0.57} \\
      {\textbf{$\text{DiT-XL/2-G (cfg $=1.50$)}^@$}} & \cellcolor[HTML]{A9DBD9}{1$\times$} & \color{blue}{\xmark}&  \cellcolor[HTML]{A9DBD9}{\textbf{2.30}} &  \cellcolor[HTML]{A9DBD9}{\textbf{4.56}} &  \cellcolor[HTML]{A9DBD9}{\textbf{276.56}} &  \cellcolor[HTML]{A9DBD9}{0.83} &  \cellcolor[HTML]{A9DBD9}{0.58} \\
      \midrule
        \bf Static - $N=2$ &  2.08$\times$ & \xmark & 2.40 & 5.15 & 269.74 & 0.82 & 0.58  \\
        \bf Static - $N=3$ & \cellcolor[HTML]{F3C88C}{2.80$\times$} & \xmark & \cellcolor[HTML]{F3C88C}{2.82} & \cellcolor[HTML]{F3C88C}{6.04} & \cellcolor[HTML]{F3C88C}{253.96} & \cellcolor[HTML]{F3C88C}{0.80} & \cellcolor[HTML]{F3C88C}{0.58}  \\
        \bf Static - $N=5$ &  4.57$\times$ & \xmark & 4.97 & 9.15 & 222.97 & 0.76  & 0.59  \\
        \bf Static - $N=7$ &  5.73$\times$ & \xmark & 9.80 & 13.62 & 180.54 & 0.68  & 0.59  \\
        \bf Static - $N=10$ &  8.07$\times$ & \xmark & 25.24	& 22.08 &  110.03& 0.53 & 0.56 \\
        \bottomrule
      \end{tabular}
      }
      \caption{\textbf{Comparative analysis of class-conditional image generation on ImageNet}. We use $\text{DiT-XL/2-G(cfg-1.50)}^@$ as our baseline model. Both the baseline and our method utilize 250 DDIM sampling steps. Our findings indicate that a static cache threshold of 3 provides an optimal balance between FID score and computational speed-up. *Results reproduced using DeepCache. @Results reproduced using FORA.}
      \label{table:main_ldm_imagenet}
      \vspace{-2mm}
  \end{table*}



\begin{table}[t]
\centering
\small
\begin{tabular}{l|c|ccc}
   \toprule
    \multicolumn{5}{c}{\bf MS-COCO 256 $\times$ 256}  \\
    \bf Method & \bf Sampler & \bf FID-30K (Inception)$\downarrow$ & \bf FID-30K (CLIP)$\downarrow$ & Speed-up$\uparrow$ \\
    \midrule
    {$\text{\model}^@$} & DPM-Solver &  \bf 28.15  & \bf 12.14 & \cellcolor[HTML]{A9DBD9}{1$\times$} \\
    {$\text{\model}^@$} & IDDPM      & 43.54 & 12.35 & \cellcolor[HTML]{A9DBD9}{1$\times$} \\
    \midrule
    FORA (Static $N$ = 2) & DPM-Solver & 38.20 & 13.91 & \cellcolor[HTML]{F3C88C}{1.54$\times$} \\
    FORA (Static $N$ = 2) & IDDPM   &  55.30 & 15.44 & \cellcolor[HTML]{F3C88C}{1.86$\times$} \\
    \bottomrule
\end{tabular}
\vspace{2mm}
\caption{\label{table:pixart} {\textbf{Comparative analysis of text-conditional image generation using T2I models}}. We evaluate our model with static caching ($N=2$) against the baseline \model, using COCO FID-30K scores (zero-shot). Hyper-parameters include a positional embedding scale of 1 for $256\times256$ image size. We employ two image features extractors: Inception V3 and CLIP-ViT-B-32 for metric evaluation. Our caching approach demonstrates accelerated sampling time while maintaining comparable FID scores. The code is sourced from GigaGAN~\citep{kang2023gigagan}. @Results reproduced using FORA. Note: While \model~\citep{chen2023pixartalpha} reports a FID score of 7.32, we omit this from the table due to lack of specificity regarding the feature extractor used and absence of reproduction code.}
\vspace{-6mm}
\end{table}

\subsection{Datasets and encoders}

To thoroughly evaluate our method, we target on two distinct types of image generation task. For class-conditional image generation, we utilized the widely-recognized ImageNet dataset~\citep{Deng2009ImageNet}, which offers a diverse range of image classes. To assess text-conditional image generation capabilities, we employed the MSCOCO dataset~\citep{Lin2014mscoco}, known for its rich textual descriptions paired with images. For the text-conditional model, we leveraged the T5 large language model~\citep{Raffel2020t5} as our text encoder, enabling robust extraction of conditional features from textual inputs.


\subsection{Evaluation models}

Our method is designed to be versatile, compatible with both unconditional and conditional models. To ensure a fair and comprehensive evaluation, we focus on conditional models, using fixed seeds for all experiments to enable equitable qualitative comparisons. Furthermore, we demonstrate that our method is agnostic to input conditions and can seamlessly integrate with state-of-the-art fast samplers. With these objectives in mind, we select two models for our evaluation: DiT~\citep{peebles2023dit} and \textsc{PixArt-}$\alpha$~\citep{chen2023pixartalpha}. Both of these models employ a Transformer backbone within the diffusion probabilistic model framework, representing the latest advancements in the field.

\vspace{-.1in}
\begin{itemize}[leftmargin=*]
    \item DiT primarily focuses on ImageNet class conditioning, showcasing the improvements achieved by using a Transformer backbone over the traditional U-Net architecture. It utilizes the DDIM fast sampler to enhance its sampling efficiency. In our experiments with DiT, we maintain the classifier-free guidance strength at 1.5, as per the original implementation.
    \item \textsc{PixArt-}$\alpha$, on the other hand, specializes in text-conditional image generation. This model is particularly noteworthy for its emphasis on reducing carbon footprint by lowering training costs while maintaining high image quality. Beyond training efficiency, \textsc{PixArt-}$\alpha$ incorporates efficient sampling algorithms such as IDDPM and DPM-Solver to further its goals. For our experiments with PixelArt-Alpha, we keep the classifier-free guidance strength at 4.5, adhering to the model's original configuration.
\end{itemize}

By evaluating FORA on the two models, we show its broad applicability and effectiveness. Our results indicate that FORA can be successfully integrated with these fast samplers, potentially reducing the carbon footprint even further by decreasing inference costs while preserving image quality.

\subsection{Experimental setup}


For the DiT model, we conduct inference using 250 time steps with DDIM, employing various thresholds for static caching. Following DiT's evaluation protocol, we sampled 50,000 images from the ImageNet dataset for our assessment. For \textsc{PixArt-}$\alpha$, we performe qualitative evaluation using sample texts provided by the model's authors. Quantitative evaluation utilized the MS-COCO-30K dataset. Our method for \textsc{PixArt-}$\alpha$ employs IDDPM sampling by default, utilizing 100 time steps, along with DPM-Solver for 20 time steps. All experiments were conducted on a single A100 GPU to ensure consistency in hardware performance.

\begin{table*}[t]
\centering
    \small
    \resizebox{\linewidth}{!}{
    \begin{tabular}{l |c c c| c c c c c }
      \toprule
      \multicolumn{8}{c}{\bf ImageNet 256 $\times$ 256}  \\
      \bf \#TimeSteps & \bf caching Interval & \bf Recomp\% & \bf origcomp & \bf FID $\downarrow$ & \bf sFID $\downarrow$ & \bf IS $\uparrow$& \bf Precision $\uparrow$& \bf Recall $\uparrow$  \\
      \midrule
       100 & 2 & 50.0 & 2800 & 4.28 & 8.61 & 231.81 & 0.77 & 0.58  \\
       100 & 3 & 33.3 & 2800 & 9.51 & 14.18 & 183.83 & 0.69 & 0.58  \\
      \cellcolor[HTML]{A9DBD9}{\bf 250} & \cellcolor[HTML]{A9DBD9}{\bf 3} & \cellcolor[HTML]{A9DBD9}{33.3} & \cellcolor[HTML]{A9DBD9}{7000} & \cellcolor[HTML]{A9DBD9}{2.82} & \cellcolor[HTML]{A9DBD9}{6.04} & \cellcolor[HTML]{A9DBD9}{253.96} & \cellcolor[HTML]{A9DBD9}{0.80} & \cellcolor[HTML]{A9DBD9}{0.58}  \\
       500 & 3 & 33.3& 14000  & 2.33 & 4.82 & 267.94 & 0.82 & 0.58  \\
       \cellcolor[HTML]{F3C88C}{\bf 500} & \cellcolor[HTML]{F3C88C}{\bf 6} & \cellcolor[HTML]{F3C88C}{16.7} & \cellcolor[HTML]{F3C88C}{14000} & \cellcolor[HTML]{F3C88C}{2.96} & \cellcolor[HTML]{F3C88C}{6.20} & \cellcolor[HTML]{F3C88C}{250.83} & \cellcolor[HTML]{F3C88C}{0.80} & \cellcolor[HTML]{F3C88C}{0.59}  \\
      \bottomrule
    \end{tabular}
    }    \caption{\label{table:ablation_timesteps_imagenet}\textbf{Ablation on the relationship between sampling time steps and caching interval for class-conditional image generation on ImageNet.} {Recomp\%} represents the ratio of reduced cached recomputations to the actual number of attention and MLP computations. Results demonstrate that as total sampling time steps increase, proportionally adjusting the caching interval yields comparable results while further reducing the number of recomputations. This study highlights the scalability and efficiency of our caching method across various sampling configurations.}
    \vspace{-2mm}
\end{table*}





\subsection{Results}{\label{para:results}}

The qualitative comparison with both DiT and \textsc{PixArt-}$\alpha$ is illustrated in Fig.~\ref{fig:teaser}.  For class-conditional image generation, our method demonstrates good quality reconstruction with a 3 to 4$\times$ speed-up. An interesting observation is that as the caching interval increases, the background gains equal importance to the class subject. While the class object is generated effectively, this opens up avenues for further research, such as exploring the significance of background features on the FID score and investigating whether masking the background can improve the score. The quantitative evaluation using the ImageNet dataset is presented in Table.~\ref{table:main_ldm_imagenet}. The results indicate that a cache interval of 3 strikes a good balance between the FID score and the speed-up.

For text-conditional image generation, qualitative results in Fig.~\ref{fig:teaser} applying FORA on top of IDDPM sampling show an artistic effect without compromising image quality. Interestingly, the DPM-Solver achieves softer image generation, enhancing realism by reducing the vibrancy present in the baseline. The quantitative results comparing our method's speed-up against the T2I model are shown in Table.~\ref{table:pixart}, highlighting the efficiency and potential aesthetic benefits of our approach.

\subsection{Ablation studies}

We conducted two ablation studies to further understand the nuances of our method.

\textbf{Caching Interval.} We investigate how adjusting the caching interval relative to the total sampling time steps minimizes recomputations while maintaining consistent evaluation metric results. The findings, presented in Table.~\ref{table:ablation_timesteps_imagenet}, show that increasing the cache interval as sampling steps increase does not negatively impact the FID or IS score. This suggests that our method can maintain image quality even with larger caching intervals, potentially allowing for even greater computational savings.

\begin{figure}[t]
    \centering
    \includegraphics[width=1\linewidth]{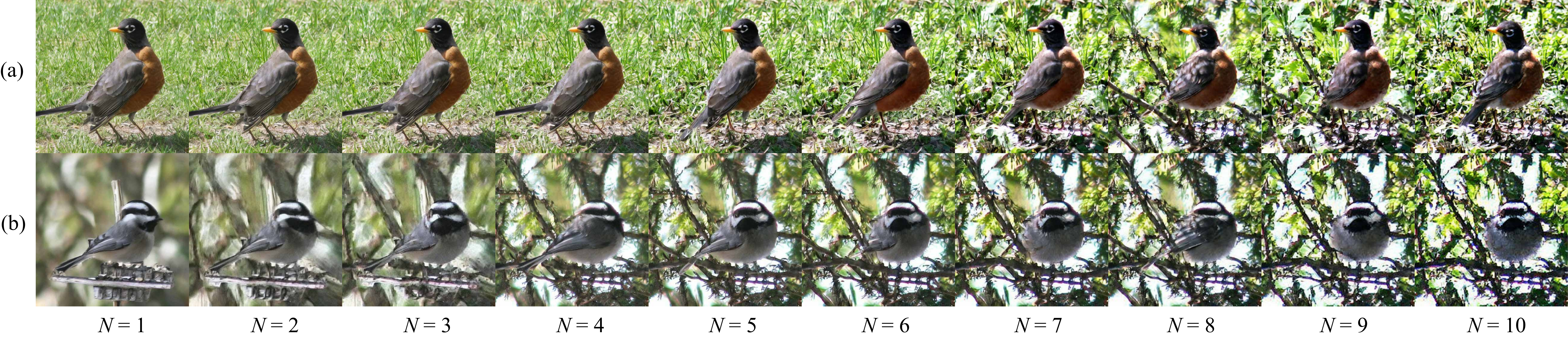}
    \caption{\label{fig:ablation_static_N_DiT}\textbf{Visual quality ablation study on caching interval $N$ for ImageNet class-conditional image generation.} This study demonstrates the progressive impact of increasing caching intervals on image quality. Results indicate that beyond an interval of 5, there is a noticeable degradation in image detail and realism. As noted in section.~\ref{para:results},  we observe that as caching intervals increase, background elements gain prominence equal to the main subject, potentially affecting the overall image composition and quality.}
\end{figure}
\begin{figure}
    \centering
    \includegraphics[width=1\linewidth]{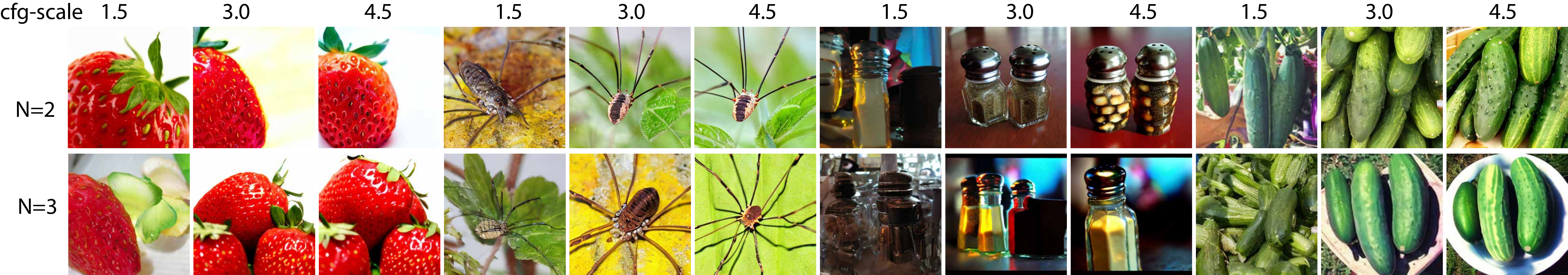}
    \caption{\label{fig:ablation_cfgscale_N_DiT}\textbf{Visual quality and composition ablation study examining the interplay between guidance strength and caching interval $N$ on ImageNet class-conditional image generation.} It reveals that a guidance strength of 1.5 frequently results in cluttered compositions with poor object-background separation and diminished overall image quality. In contrast, higher guidance strengths demonstrate improved image clarity and composition.}
\end{figure}

\textbf{Conditional Guidance Strength.} We analyze the effect of guidance strength on image quality and diversity. Table.~\ref{table:ablation_cfgscale_imagenet}  highlights the importance of guidance strength for better image quality. The qualitative results are shown in Fig.~\ref{fig:ablation_cfgscale_N_DiT}. Interestingly, we find that both the guidance strength and the cache interval affect the composition of the image. As these parameters change, we observe that the focus of the subject can become diluted, with equal importance given to the background, or the image may contain several superimposed objects.

\begin{table*}[t]
\centering
    \small
    \begin{tabular}{l | c c c c c }
      \toprule
      \multicolumn{6}{c}{\bf ImageNet 256 $\times$ 256 (250 sampling time steps)}  \\
      \bf cfg\-scale & \bf FID $\downarrow$ & \bf sFID $\downarrow$ & \bf IS $\uparrow$& \bf Precision $\uparrow$& \bf Recall $\uparrow$  \\
      \midrule
       1.5 & \bf 2.82 & \bf 6.04 & 253.96 & 0.80 & \bf 0.58  \\
       3.0 & 12.58 & 6.79 & 451.28 & \bf 0.93 & 0.35  \\
       4.5 & 17.19 & 10.91 & \bf 483.50 & \bf 0.93 & 0.24  \\
      \bottomrule
    \end{tabular}
    \caption{\label{table:ablation_cfgscale_imagenet}\textbf{Ablation on cfg-scale for class-conditional image generation on ImageNet, focusing on Inception Score (IS) improvements.}  Results demonstrate a positive correlation between increasing guidance scale and higher IS, indicating enhanced image quality and diversity.}
    \vspace{-2mm}
\end{table*}

\section{Conclusion}

We present a simple yet effective caching mechanism applicable to any transformer-based diffusion Model. Our approach significantly reduces computational overhead, thereby accelerating the sampling and inference process. The proposed model is a training-free, plug-and-play solution that can be seamlessly integrated with fast samplers. Through extensive experimentation and ablation studies, we have demonstrated the efficiency and versatility of our method.

One limitation of our current approach is that the static caching mechanism cannot fully utilize the complex similarity patterns inherent in transformer-based diffusion models. These models often exhibit dynamic and context-dependent similarities across different stages of the diffusion process, which static caching may not optimally capture. A more sophisticated dynamic caching mechanism, capable of adapting to these evolving patterns in real-time, is desirable for future development. Such an approach could potentially lead to even greater efficiency gains and better preservation of image quality across various sampling scenarios.

\interlinepenalty=10000  
\bibliographystyle{plainnat}
\bibliography{ref}

\end{document}